\RequirePackage{fix-cm}
\documentclass[twocolumn]{svjour3}          % twocolumn
\smartqed  % flush right qed marks, e.g. at end of proof
\usepackage{graphicx}
\usepackage{subfigure}
\usepackage{amsmath}
\usepackage{graphicx}
\usepackage{algorithm}
\usepackage{algorithmicx}
\usepackage{algpseudocode}
\usepackage{bm}
\usepackage{makecell}
\usepackage{diagbox}
\usepackage{indentfirst}
\usepackage[misc]{ifsym}
\usepackage{amsfonts}
\usepackage[table]{xcolor}
\usepackage{multirow}
\usepackage{makecell}
\usepackage{booktabs}

\usepackage{picins}

\renewcommand{\algorithmicrequire}{\textbf{Input:}}  % Use Input in the format of Algorithm
\renewcommand\tabcolsep{3.0pt}

\begin{document}
	
	\title{SONAR: Semantic-Object Navigation with Aggregated Reasoning through a Cross-Modal Inference Paradigm%\thanks{Grants or other notes
	}
	\subtitle{}
	
	
	\author{Yao Wang$^{1,2}$ \and Zhirui Sun$^1$ \and Wenzheng Chi$^3$ \and Baozhi Jia$^{4,5}$ \and Wenjun Xu$^2$ \and Jiankun Wang$^{1,}$*
	}

	
        \institute{
        \Letter~Jiankun Wang \\
        \email{wangjk@sustech.edu.cn} \\
        * Corresponding author \\
        \\
        $^1$ Shenzhen Key Laboratory of Robotics Perception and Intelligence, 
        Department of Electronic and Electrical Engineering, 
        Southern University of Science and Technology, Shenzhen, China. \\
        $^2$ Research Institute of Multiple Agents and Embodied Intelligence, 
        Peng Cheng Laboratory, Shenzhen, China.\\
        $^3$ Robotics and Microsystems Center, School of Mechanical and Electric Engineering, Soochow University, Suzhou, China.\\
        $^4$ Xiamen Key Laboratory of Visual Perception Technology and Application, Xiamen, China.\\
        $^5$ Algorithm Research Center, Reconova Information Technology Co., Ltd., Xiamen, China.
        }

	\date{Received: date / Accepted: date}

	\maketitle

\begin{abstract}
% 理解人类指令并在未知环境中完成视觉 - 语言导航任务对机器人至关重要。然而，现有的模块化方法严重依赖训练数据的质量，且泛化能力往往较差。基于视觉 - 语言模型的方法虽然具有很强的泛化能力，但在语义线索较弱时表现往往不尽人意。为解决这些问题，本文提出了 SONAR，一种通过跨模态范式的聚合推理方法。该方法将基于语义地图的目标预测模块与基于视觉 - 语言模型的值地图模块相结合，使得在具有不同语义线索水平的未知环境中能够进行更稳健的导航，并有效地平衡了泛化能力与场景适应性。在目标定位方面，我们提出了一种将多尺度语义地图与置信度地图相结合的策略，旨在减少目标物体的误检测。
% 我们在 Gazebo 模拟器中对 SONAR 进行了评估，利用最具挑战性的 Matterport 3D（MP3D）数据集作为实验基准。实验结果表明，SONAR 的成功率达到 38.4%，SPL 为 17.7%。
Understanding human instructions and accomplishing Vision-Language Navigation tasks in unknown environments is essential for robots. However, existing modular approaches heavily rely on the quality of training data and often exhibit poor generalization. Vision-Language Model based methods, while demonstrating strong generalization capabilities, tend to perform unsatisfactorily when semantic cues are weak. To address these issues, this paper proposes SONAR, an aggregated reasoning approach through a cross modal paradigm. The proposed method integrates a semantic map based target prediction module with a Vision-Language Model based value map module, enabling more robust navigation in unknown environments with varying levels of semantic cues, and effectively balancing generalization ability with scene adaptability. In terms of target localization, we propose a strategy that integrates multi-scale semantic maps with confidence maps, aiming to mitigate false detections of target objects.
We conducted an evaluation of the SONAR within the Gazebo simulator, leveraging the most challenging Matterport 3D (MP3D) dataset as the experimental benchmark. Experimental results demonstrate that SONAR achieves a success rate of 38.4\% and an SPL of 17.7\%.
\end{abstract}
	
\section{Introduction}
\label{intro} 
In an unknown environment, for a robot to accurately understand human instructions and complete vision language navigation tasks, it needs to rely on limited visual and linguistic cues to develop efficient exploration strategies while achieving precise identification of target objects\cite{batra2020objectnav}. The correlation between known objects in the environment and the target object can serve as a key guide during this process. For instance, when a robot receives the instruction `find a toilet', it does not need to search aimlessly. Instead, it can recognize objects strongly associated with bathroom scenarios, such as beds and sinks, and infer that the target object is most likely located in the nearby areas where these objects appear. This reasoning helps reduce the exploration range and improves navigation efficiency.
% \begin{figure}[t]
% \centering
% \includegraphics[width=3.3in]{wangyao/images/1.png}
% \caption{aaa}
% \label{fig1}
% \end{figure}

The core advantage of end-to-end navigation methods \cite{mirowski2016learning,codevilla2018end,singh2023scene,ramrakhya2022habitat} lies in their streamlined workflow. This workflow enables direct end-to-end learning of the entire navigation system without the need for additional map construction and maintenance. Meanwhile, current research has conducted in-depth exploration around visual representation and policy learning. It provides theoretical support for the performance optimization of this method. However, their limitations are also prominent. On the one hand, the learning difficulty is relatively high. They require the simultaneous mastery of three core capabilities—localization, map construction, and path planning. On the other hand, there is a significant simulation to reality transfer gap. The observation information in the simulated environment often deviates from that in real-world scenarios due to the constraints of simulator fidelity. In addition, the end-to-end learning paradigm based on reinforcement learning or imitation learning imposes high demands on computational resources and time consumption. All these factors collectively restrict their promotion and application in practical scenarios.
In contrast, modular learning methods\cite{chaplot2020object,hahn2021no,chaplot2020neural} integrate the dual advantages of learning based methods and traditional navigation processes. This integration allows them to decompose the target object navigation task into three core components: the mapping module, the long-term target policy module, and the path planning module. Among these components, the mapping module can provide environmental representation. This representation enables effective decoupling between the agent perception module and the subsequent policy module as well as path planning module. This feature makes modular methods easier to transfer to real-world scenarios. At the same time, they reduce the demand for computational resources and time to a certain extent. This alleviates the resource consumption problem of end-to-end methods. Yet their limitations are mainly manifested in two aspects. One is that the design of mapping modules varies significantly across most modular methods. The other is that the construction of corresponding policies relies on different premise assumptions. They lack the unified framework that end-to-end methods possess. This lack makes it difficult to conduct horizontal comparative analysis on the effectiveness of different mapping modules. It also restricts the performance evaluation and improvement among methods.
Meanwhile, Zero-shot methods\cite{yokoyama2024vlfm,yin2024sg,yu2023l3mvn} have the core advantage of strong generalization ability: without relying on a large amount of annotated data for specific targets, they can quickly adapt to new categories and new scenarios, effectively breaking the strong dependence of traditional methods on annotated data and providing a feasible path for handling navigation tasks with numerous categories and scarce data in reality. However, their limitations are also obvious: when the semantic cues in the scenario are weak, it is difficult for them to establish reliable feature-category associations, leading to a significant decline in task performance and often poor results.

Based on the above analysis, this paper proposes a novel robot navigation method namely SONAR. This method has a dynamically adaptive exploration mechanism. When the intensity of semantic cues in the environment is high, the robot focuses its exploration on the current observation area. This helps make full use of local semantic information and improve exploration efficiency. When the intensity of semantic cues is low, the robot predicts the potential position of the target object by invoking the global semantic map and then plans the exploration path.
To accurately select the optimal exploration boundary point, this study proposes a Dual-model Aggregation Reasoning (DAR) method. The specific implementation process is as follows. First, the prediction points output by the target prediction model are converted into a Distance Map. At the same time, the confidence scores output by the Vision-Language Model are mapped into a Value Map. Subsequently, the final DAR score is calculated by aggregating reasoning weights. Finally, the target point with the highest DAR score is selected as the core target point for the navigation task. This ensures the effectiveness and accuracy of exploration decisions.
In addition, to avoid the misjudgment of the target object caused by the one-sidedness of local information, this method innovatively integrates multi-scale semantic maps and confidence maps. Through the complementation of multi-source information, it achieves accurate inference of the target object position and effectively reduces the impulsive error in the robot decision making.

Our main contributions are as follows:
\begin{itemize}
    \item 
    % 我们提出了SONAR，一种可以聚合跨模态信息指导机器人完成视觉语言导航任务新的导航方法
    We propose SONAR, a novel navigation method that can aggregate cross-modal information to provide accurate guidance for robots, enabling them to efficiently accomplish Vision-Language navigation tasks.
    \item 
    % 我们提出了语义线索强度计算方法，该方法能够有效助力机器人，无论所处环境的语义线索强度高低，均能精准选取最优探索边界点 
    We propose a Semantic Cue Intensity Calculation method, which can effectively assist robots in accurately selecting the optimal exploration boundary points regardless of the intensity of semantic cues in the environment.
    \item 
    % 我们提出了多尺度语义地图和置信度地图的融合方法，帮助机器人更准确的定位目标物体位置 
    We propose a fusion method of multi-scale semantic maps and confidence maps to help robots locate target objects more accurately.
\end{itemize}

\section{Related Work}
\label{sec:related_work}

Object Goal Navigation (ObjectNav) task has attracted increasing attention in embodied AI and robotics. It requires an agent to perceive its environment, reason about spatial-semantic relationships, and plan actions to reach a target object. Existing research can be broadly categorized into three paradigms. The first is End-to-End methods. These methods directly map sensory inputs to navigation actions. The second is modular approaches. These approaches decompose the task into perception, mapping, and planning components. The third is Zero-shot methods. These methods focus on generalization to novel objects and instructions without task-specific training. We review these three lines of research as follows.
\subsection{End-to-End Object Navigation}
End-to-End approaches address ObjectNav by learning a direct mapping from sensory observations to actions, typically via reinforcement learning (RL) or imitation learning (IL) \cite{liu2020indoor}. In these methods, the agent encodes visual inputs and integrates target information and previous actions to guide the policy network, which is trained through environment interactions. Research efforts have focused on improving visual representations, including raw RGB-D images, object detections, segmentation masks, as well as relational embeddings that encode object-object and object-region relationships \cite{du2020learning,druon2020visual,campari2020exploiting,pal2021learning,zhang2021hierarchical,mayo2021visual,fukushima2022object}. In addition, significant attention has been paid to enhancing policy learning to address challenges such as sparse rewards, overfitting, and low sample efficiency. Advanced strategies include meta learning\cite{wortsman2019learning,li2021multi,li2020unsupervised}, hierarchical decomposition\cite{dang2023search,dang2023multiple,staroverov2023skill,wang2023skill}, auxiliary tasks\cite{ye2021auxiliary}, forward models\cite{moghaddam2022foresi,zhang2023layout}, and human-in-the-loop guidance\cite{singh2022ask4help}. While End-to-End pipelines eliminate the need for explicit maps and offer conceptual simplicity, they face challenges in real-world deployment due to high computational demands, sim-to-real transfer gaps, and the difficulty of jointly learning perception, localization, and navigation.

\subsection{Modular Object Navigation}
Modular approaches decompose ObjectNav into sequential sub tasks, such as mapping, localization, semantic reasoning, and policy planning. By explicitly modeling intermediate representations like semantic maps or value maps, modular methods leverage structured priors to improve interpretability and generalization. This design allows each module to be optimized independently and enables the integration of domain knowledge, making them more robust in long-horizon navigation. 
For example, the mapping module of SemEXP \cite{chaplot2020object} generates a global semantic map using MaskRCNN \cite{he2017mask}, while its policy module predicts long-term goals.
SSCNav \cite{liang2021sscnav} uses an encoder-decoder network to complete the local map and add semantic labels, so as to generate the next action.
Nonetheless, modular pipelines may accumulate errors across stages and require careful design of information interfaces to avoid performance degradation.

\subsection{Zero-shot Object Navigation}
Zero-shot ObjectNav aims to navigate towards novel objects or categories that are unseen during training, without requiring task specific fine tuning. Recent works leverage large scale Vision-Language Models or pre-trained multimodal embeddings to align semantic concepts with spatial observations. Such approaches benefit from knowledge transfer across modalities and tasks, enabling robots to interpret high level instructions and generalize to open vocabulary settings.
For example, VLFM\cite{yokoyama2024vlfm} utilizes the pre-trained BLIP-2 Vision-Language Model to construct a value map and combines it with a frontier map to generate navigation waypoints.
Despite these advantages, Zero-shot methods remain challenged by domain shifts between training datasets and embodied environments, as well as the difficulty of grounding abstract semantics into actionable navigation strategies.

\begin{figure*}[t]
\centering
\includegraphics[width=6.5in]{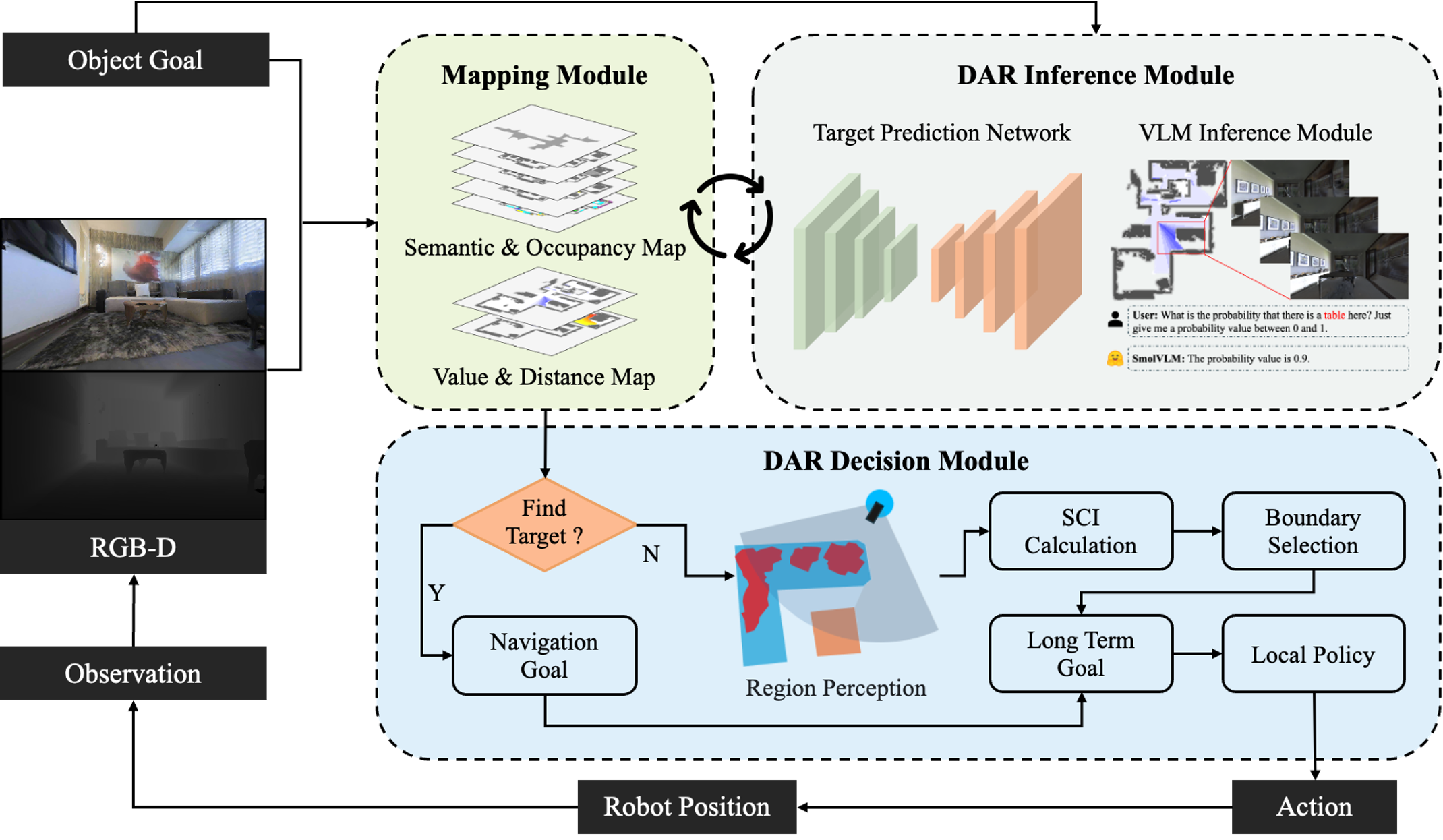}
\caption{Overview of the proposed SONAR algorithm. SONAR takes the target object and RGB-D inputs to generate semantic and occupancy maps, which are further fused with value and distance maps from the DAR Inference module to construct a unified environment representation. Based on this representation, the DAR Decision module navigates the agent to a detected target via a local policy, or, when no target is detected, adaptively allocates decision weights according to semantic cues and establishes long-term exploratory goals to guide the search process.}
\label{framework}
\end{figure*}

\section{Methodology}
\label{sec:methods}
We propose a modular paradigm called SONAR to tackle the ObjectNav task. The overview of SONAR is shown in Fig. \ref{framework}. It consists of a learnable module and a pre-trained module, namely the ‘Target Prediction Module’ and the ‘Vision-Language Model Module’. The Target Prediction Module predicts the potential location of the target based on the real-time constructed semantic map.
The Vision-Language Model Module provides the probability that the target object exists in the current region based on real-time observed images.
During the exploration phase, the robot simultaneously constructs a single-target semantic map and a multi-object semantic map. At the same time, it continuously updates the frontier boundary map, value map, and predicted target map, and selects the frontier boundary point with the highest value as the exploration target point.
When the target object appears in the semantic map, the final navigation target position is locked from the fusion map of the single-target semantic map and the multi-object semantic map.
\subsection{Multi-scale Fusion Map Builder}
% 多尺度语义融合地图构建器是 SONAR 范式的核心空间和语义感知基础。它整合多源感知信息，构建一组相互关联的多维地图层。这些地图层共同为机器人探索决策和目标定位提供全面的环境理解，涵盖语义属性、空间占用情况、探索状态以及物体的置信度。具体而言，多尺度语义融合地图由以下关键地图组件组成：
The Multi-scale Fusion Map Builder serves as the core spatial and semantic perception foundation of the SONAR paradigm. It integrates multi-source perceptual information to construct a set of interrelated, multi-dimensional map layers. These layers together provide comprehensive environmental understanding for robot exploration decision-making and target localization, covering semantic attributes, spatial occupancy, exploration status, and confidence levels of objects. Specifically, the multi-scale semantic fusion map consists of the following key map components:

\textbf{Semantic-oriented Map Layers:} These layers focus on capturing the semantic information of objects in the environment, distinguishing between the target object of interest and other non-target objects to support precise target localization.

\textbf{1) Single-Target Semantic Map:} 
A dedicated semantic map layer is maintained for the specific navigation target, denoted as $\text{SMap}_{\text{target}}$. 
Each cell in this layer stores a binary label indicating target presence or absence, where 1 corresponds to presence and 0 corresponds to absence. 
As agent acquires new visual observations, semantic recognition module detects target object and estimates its spatial coordinates within current field of view. 
These detections are projected onto $\text{SMap}_{\text{target}}$, where the corresponding cells are updated accordingly. 
Consequently, $\text{SMap}_{\text{target}}$ provides a continuously refined record of all regions in which the target has been observed, enabling reliable tracking of its potential locations throughout the navigation task.

\begin{figure}[t]
\centering
\includegraphics[width=3.3in]{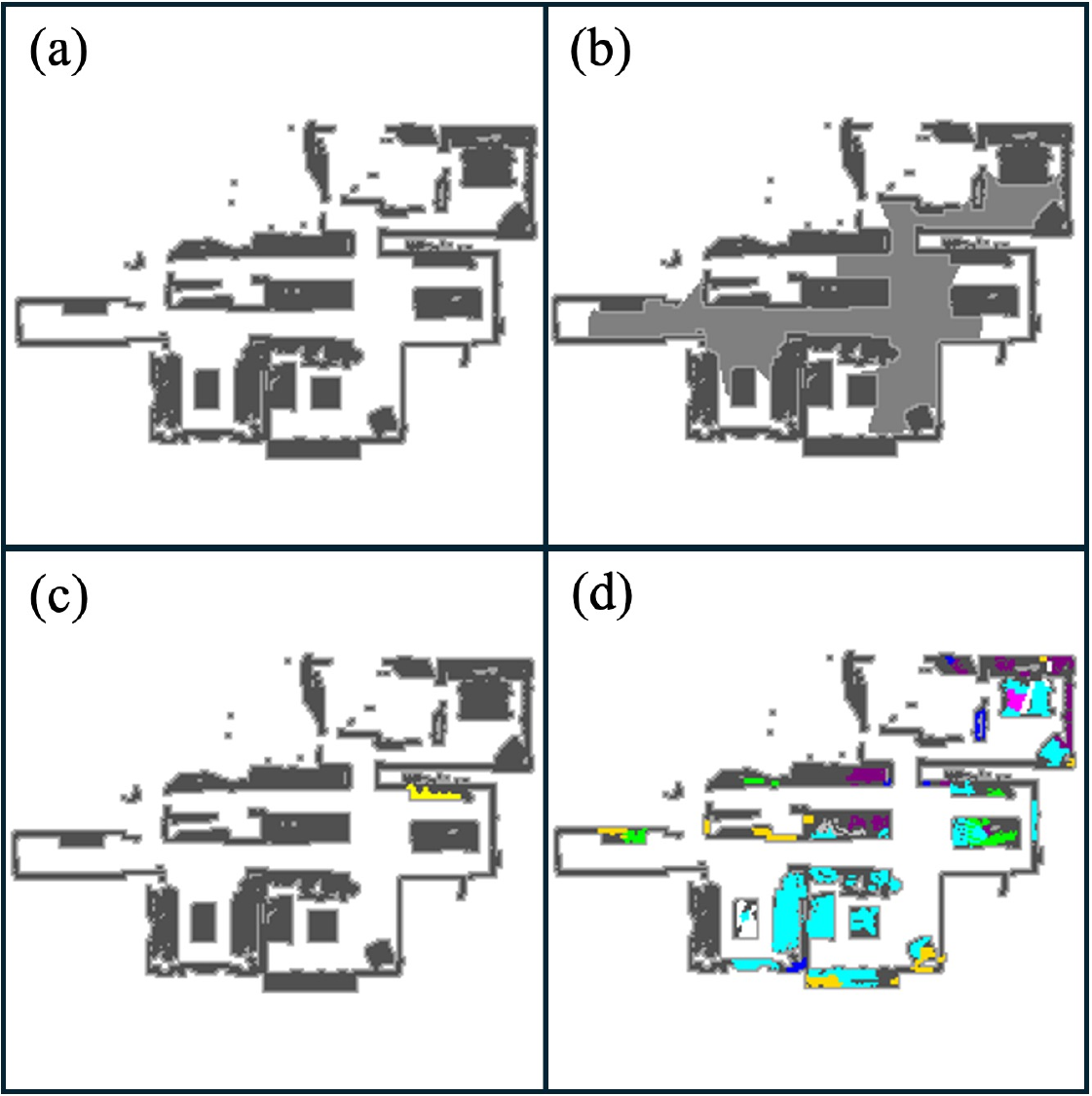}
\caption{Fusion Map. (a) Obstacle Map; (b) Explored Map; (c) Single-Target Semantic Map; (d) Multi-Object Semantic Map}
\label{fusion map}
\end{figure}

\textbf{2) Multi-Object Semantic Map:} 
A semantic map layer, denoted as $\text{SMap}{\text{multi}}$, is maintained to represent all detected object categories in the environment. 
Unlike $\text{SMap}{\text{target}}$, which focuses on a single object of interest, $\text{SMap}{\text{multi}}$ represents multiple semantic categories by assigning class-specific labels to each cell (e.g., 1 for table, 2 for sofa).
As the agent navigates and acquires new observations, detected objects are continuously projected into $\text{SMap}_{\text{multi}}$ according to their spatial coordinates, thereby refining the map over time. 
This process allows $\text{SMap}_{\text{multi}}$ to preserve a global semantic layout of the scene. 
Such contextual knowledge not only improves the understanding of the surrounding environment, but also facilitates higher-level reasoning for navigation, such as inferring the likely proximity of a target chair to nearby tables or predicting traversable regions around walls and large furniture.

\textbf{Confidence-oriented Map Layers:} These layers quantify the reliability of semantic recognition results. Their core function is to reduce the risk of erroneous decisions caused by semantic information errors in subsequent decision-making processes by modeling and addressing uncertainties in visual perception.

\textbf{1) Single-Target Confidence Map:}
A confidence map layer $\text{CMap}_{\text{target}}$ corresponding to the single-target semantic map $\text{SMap}_{\text{target}}$ is maintained to record confidence levels of target detections. Each cell in $\text{CMap}_{\text{target}}$ stores the confidence of observing the target at that location. As new observations are acquired, the current confidence value at each relevant cell is retrieved. If the new detection confidence exceeds the existing value, $\text{SMap}_{\text{target}}$ is updated with the target label and $\text{CMap}_{\text{target}}$ is updated with the new confidence. If the new confidence is lower or equal to the existing value, the confidence is updated by averaging the existing and new confidence. This update mechanism ensures that $\text{SMap}_{\text{target}}$ maintains the most reliable record of observed target locations while continuously integrating new observations and confidence information.
\begin{equation}
\text{CMap}_{\text{target}} =
\begin{cases} 
C, & C \ge \text{CMap}_{\text{target}} \\[1mm]
\dfrac{\text{CMap}_{\text{target}} + C}{2}, & C < \text{CMap}_{\text{target}} 
\end{cases}
\end{equation}
herein, $\text{CMap}_{\text{target}}$ denotes the corresponding confidence map and $C$ indicates the observed confidence of the target.

\begin{algorithm}[t]
\renewcommand{\algorithmicrequire}{\textbf{Input:}} % 将REQUIRE显示为Input
\caption{Multi-scale Semantic Fusion Map Builder}
\label{alg:multi_scale_map}
\begin{algorithmic}[1]
\Require Object List $L$, Target Object $T$, Detected Object $D$, Detection Confidence $C$, Robot Position $P$

\State $\text{SMap}_{\text{target}} \gets \emptyset$; $\text{SMap}_{\text{multi}} \gets \emptyset$
\State $\text{CMap}_{\text{target}} \gets \emptyset$; $\text{CMap}_{\text{multi}} \gets \emptyset$
\State $\mathbf{O} \gets \emptyset$; $\mathbf{E} \gets \emptyset$; $\mathbf{F} \gets \emptyset$
\While{Not STOP}
    \State $D \gets \text{GetObservation()}$
    \For{$D \in L$}
        \State $\text{CMap}_{\text{multi}} \gets \text{UpdateConfidenceMap($C$)}$
        \State $\text{SMap}_{\text{multi}} \gets \text{UpdateSemanticMap($D,C,CMap$)}$

        \If{$D$ matches target $T$}
            \State $\text{CMap}_{\text{target}} \gets \text{UpdateConfidenceMap($C$)}$
            \State $\text{SMap}_{\text{target}} \gets \text{UpdateSemanticMap($C$)}$
        \EndIf
    \EndFor
    \State$\mathbf{O} \gets$ ExtractObstacles($D$)
    \State$\mathbf{E} \gets$ UpdateExploredRegion($P,O$)
    \State$\mathbf{F} \gets$ ComputeFrontiers($O,E$)
\EndWhile
\State \textbf{return} $SMap, CMap$
\end{algorithmic}
\end{algorithm}

\textbf{2) Multi-Object Confidence Map:}
A multi-object confidence map $\text{CMap}_{\textbf{multi}}$ corresponding to the multi-object semantic map $\text{SMap}_{\text{multi}}$ is maintained to record confidence levels of detected object categories. Each cell in $\text{CMap}_{\text{multi}}$ stores the confidence of observing a particular object class at that location. As new observations are acquired, for each pixel corresponding to a detected object, the current confidence value is retrieved. If the new detection confidence exceeds the existing value, $\text{SMap}_{\text{multi}}$ is updated with the corresponding object class label and $\text{CMap}_{\text{multi}}$ is updated with the new confidence. If the new confidence is lower than the existing value and the current semantic label matches the detected object class, the confidence value in $\text{CMap}_{\text{multi}}$ is updated by averaging the existing and new confidence. This update mechanism ensures that $\text{SMap}_{\text{multi}}$ maintains a reliable global semantic layout of the environment while continuously integrating new multi-object observations and their confidence information.
\begin{equation}
\text{CMap}_{\text{multi}} =
\begin{cases} 
C, & C \ge \text{CMap}_{\text{multi}} \\[1mm]
\dfrac{\text{CMap}_{\text{multi}} + C}{2}, & C < \text{CMap}_{\text{multi}} 
\end{cases}
\end{equation}

\begin{equation}
\text{SMap}_{\text{multi}} =
\begin{cases} 
l_{\text{obj}}, &  C > \text{CMap}_{\text{multi}} \\[1mm]
\text{SMap}_{\text{multi}}, & \text{otherwise}
\end{cases}
\end{equation}
herein, $\text{SMap}_{\text{multi}}$ denotes the multi-object semantic map, $\text{CMap}_{\text{multi}}$ denotes the corresponding confidence map, $C$ indicates the observed confidence of a detected object, and $l_{\text{obj}}$ represents the object class label.

\begin{algorithm}[t]
\renewcommand{\algorithmicrequire}{\textbf{Input:}} % 将REQUIRE显示为Input
\caption{VLM-based Value Map Update}
\label{alg:value_map_update}
\begin{algorithmic}[1]
\Require Robot Position $P$, Robot Heading $\phi$, Field of View $\theta$, Obstacle Map $\mathbf{O}$
\State $T \gets \text{GetPrompt()}$
\State $I \gets \text{GetObservation()}$
\State $S \gets \text{ReasonSmolVLM}(I, T)$

\For{each angle $\alpha \in [\phi - \theta/2, \phi + \theta/2]$ }
    \State $S_w \gets \text{WeightedScore}(\alpha, \phi, S)$
    \For{distance $d \in[0,d_{\max}]$}
        \State $C \gets \text{PolarToCartesian}(P,\alpha,d)$
        \If{$C \in \mathbf{O}$}
            \State \textbf{terminate current ray}
            \State \textbf{break}
        \Else
            \State $V(C) \gets \text{UpdateValue}(V(C), S_w)$
        \EndIf
    \EndFor
\EndFor

\State \textbf{return} $V$
\end{algorithmic}
\end{algorithm}

\textbf{Occupancy-oriented Map Layers:} 
These layers encode the physical constraints of the environment as well as the progress of the robot during exploration. 
They provide guidance for selecting safe and informative directions of movement, ensuring both collision avoidance and efficient coverage of unexplored regions.

\textbf{1) Obstacle Map:} 
A binary occupancy map $\mathbf{O}$ is maintained to indicate regions inaccessible to the robot and traversable regions, where 1 corresponds to occupied and 0 corresponds to free.
This map is constructed using depth sensor measurements: when a physical obstacle is detected at a specific coordinate, the corresponding cell in the map is set to 1. 
The obstacle map serves as a fundamental component for safe navigation, as both global and local planners strictly avoid trajectories that intersect with occupied regions, thereby preventing collisions.

\begin{equation}
\mathbf{O}(x,y) =
\begin{cases}
1, & \text{Occupied} \\
0, & \text{Free}
\end{cases}
\end{equation}

\textbf{2) Explored Region Map:}
A binary map $\mathbf{E}$ is maintained to record the exploration coverage of the robot. 
Cells corresponding to regions that have been visually observed and unobserved areas are marked with binary labels, where 1 corresponds to explored and 0 corresponds to unexplored.
As the robot navigates, the regions within its field of view are continuously updated as explored. 
This map enables the agent to avoid redundant exploration and prioritize unexplored areas.
\begin{equation}
\mathbf{E}(x,y) =
\begin{cases}
1, & \text{Observed} \\
0, & \text{Unobserved}
\end{cases}
\end{equation}

\textbf{3) Frontier Boundary Map:} 
A map $\mathbf{F}$ is generated by combining the Explored Region Map and the Obstacle Map. Cells corresponding to locations adjacent to unexplored regions and free of obstacles are marked as frontier boundary points. These points represent the edge of the robot current perceptual range. The robot evaluates the value of each frontier boundary point and prioritizes the point with the highest value to
efficiently expand the explored area.
\begin{equation}
\mathbf{F}(x,y) = 
\begin{cases}
1, & \text{if } \mathbf{E}(x,y)=1, \\ 
   & \text{and } \mathbf{O}(x,y)=0, \\
   & \text{and } \exists (x',y') \in \mathcal{N}(x,y): \mathbf{E}(x',y')=0, \\
0, & \text{otherwise}
\end{cases}
\end{equation}
where \(\mathcal{N}(x,y)\) denotes the 8-connected neighborhood of cell \((x,y)\).

% Each frontier point is assigned a utility value $V(x,y)$, and the next exploration target is selected as:  

% \begin{equation}
% (x^*,y^*) = \arg\max_{(x,y)\in\mathbf{F}} V(x,y)
% \end{equation}

\subsection{VLM based Value Map Module}
% 价值地图模块是本方法中的核心决策依据，其作用是评估环境中各个区域与目标物体的相关性，并以此指导探索过程中的前沿选择。与传统仅依赖空间边界的前沿地图不同，价值地图在二维自顶向下的网格中引入了语义信息，从而使机器人能够在探索时兼顾“覆盖效率”和“目标导向”。

% 在实现上，价值地图仅包含一个\emph{语义价值通道}，用于存储每个像素点在定位目标物体时的重要程度。为了获得该语义价值，我们采用了轻量化的预训练视觉语言模型 **SmolVLM**。该模型能够直接将图像与文本提示进行对齐，从而输出区域与目标物体之间的相关分数。具体而言，机器人输入当前的 RGB 图像和描述目标的文本提示（如“前方可能有一个\textit{<目标物体>}”），SmolVLM 会计算图像特征与文本特征之间的相似度。该分数越高，表示当前观测区域越可能包含目标物体。随后，这些分数会被投影到自顶向下的价值地图中，形成对环境的语义化刻画。

% 随着机器人不断探索，价值地图会在新的视觉观测驱动下持续更新，从而逐步形成全局的语义价值分布。基于这一分布，机器人能够在所有候选前沿区域中挑选出价值最高的位置作为下一个探索目标，实现面向任务的高效导航。

The Value Map Module serves as the core decision making component in our method, evaluating the relevance of each region in the environment to the target object and guiding frontier selection during exploration. Unlike traditional frontier maps that depend solely on spatial boundaries, the value map incorporates semantic information in a top-down 2D grid, allowing the robot to balance exploration efficiency and task oriented navigation.

Formally, the value map $\mathbf{V}$ is defined as a 2D grid with a single channel, where each cell $(x,y)$ represents the semantic importance of that location for localizing the target object:
\begin{equation}
\mathbf{V}(x,y) \in \mathbb{R}, \quad (x,y) \in \Omega
\end{equation}

During exploration, the robot computes a semantic score $S$ for the observed region using a lightweight pre-trained Vision-Language Model \textbf{SmolVLM}. Given the current RGB image $I$ and a textual prompt describing the target $T$, SmolVLM outputs a similarity score:
\begin{equation}
S = \text{SmolVLM}(I, T)
\end{equation}
where $I$ is the raw RGB image of the observed region, $T$ is the query prompt, and $\text{SmolVLM}(\cdot)$ takes $I$ and $T$ as inputs, processes them via internal Vision-Language reasoning, and outputs $S$ as the predicted likelihood of the target object being present.

The value map is updated using a field of view sector filling method. Assume the robot is located at $(x_r, y_r)$ with horizontal FOV $\theta_{\text{fov}}$ and heading $\phi_r$. Each ray is cast along angles and points along the ray are mapped to grid coordinates:
\begin{equation}
\alpha \in [\phi_r - \theta_{\text{fov}}/2, \phi_r + \theta_{\text{fov}}/2]
\end{equation}
\begin{equation}
(x, y) = (x_r + d \cdot \cos\alpha, \; y_r - d \cdot \sin\alpha), \quad d \in [0, d_{\max}]
\end{equation}
where $d_{\max}$ is the maximum observable distance. The ray stops when it encounters an obstacle.

For each reachable cell $(x,y)$, the value map is updated by averaging the previous value $S$ with the newly observed semantic score:
\begin{equation}
\mathbf{V}(x,y) \gets \frac{\mathbf{V}(x,y)^2 + S^2}{\mathbf{V}(x,y) + S}
\end{equation}

Through this continuous update mechanism, the value map gradually forms a global semantic distribution, enabling the robot to select the frontier with the highest value as the next exploration target and achieve efficient task oriented navigation.

\begin{figure}[t]
\centering
\includegraphics[width=3.3in]{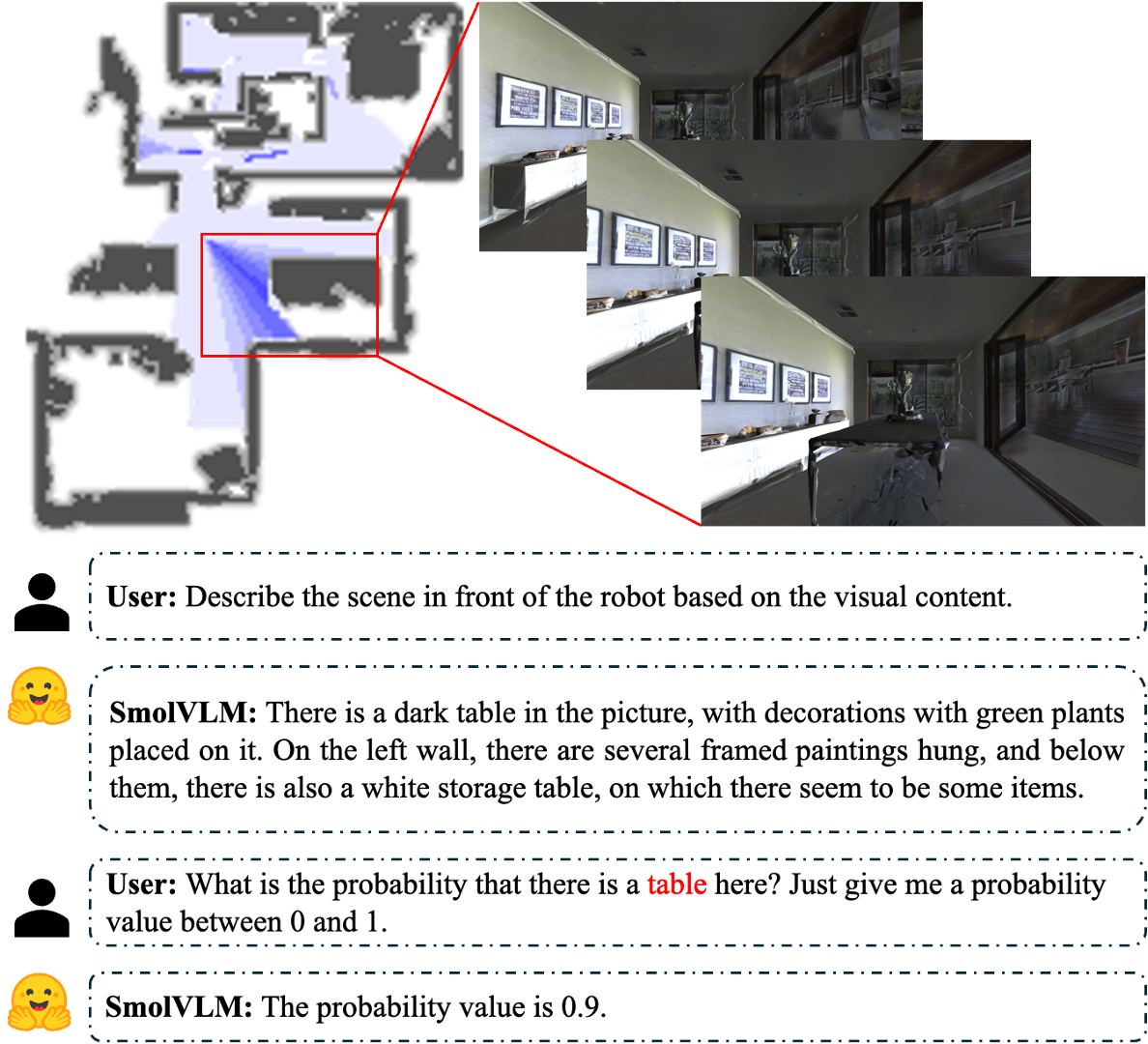}
\caption{Region Reasoning with VLM. The value map indicates the semantic importance of each cell for the target object. RGB observations are processed by SmolVLM, and scores are projected along the robot field of view to update the map.}
\label{value map}
\end{figure}

% \begin{figure}[t]
% \centering
% \includegraphics[width=3.3in]{images/5_distance map.png}
% \caption{The predicted distance map. A Local Semantic Map is processed by the network to produce a Predicted Semantic Map highlighting the target. The Predicted Distance Map encodes the distance from the predicted target, with blue representing closer regions and red representing farther regions.}
% \label{distance map}
% \end{figure}

\begin{figure}[t]
\centering
\includegraphics[width=3.3in]{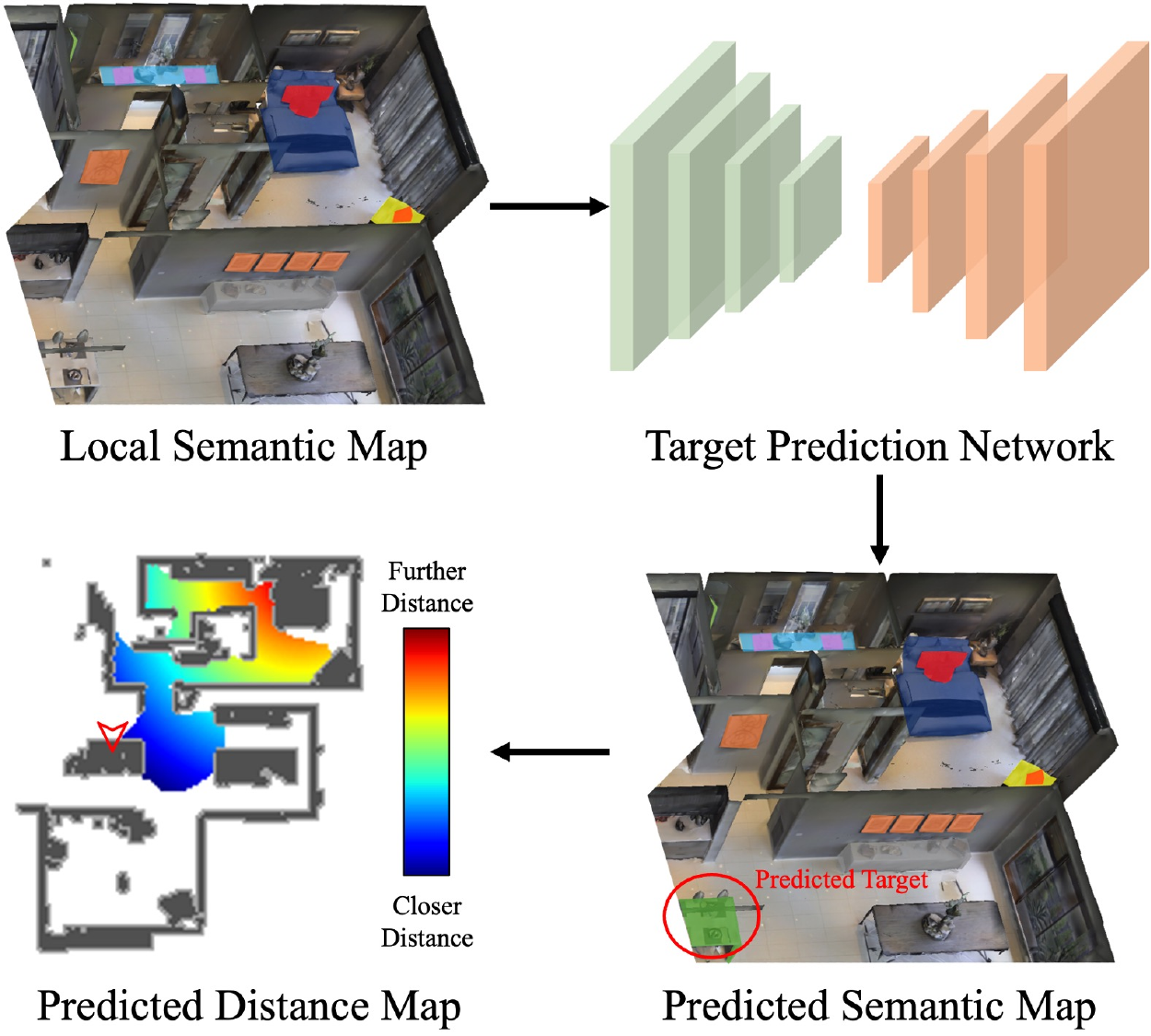}
\caption{The predicted distance map. A Local Semantic Map is processed by the network to produce a Predicted Semantic Map highlighting the target. The Predicted Distance Map encodes the distance from the predicted target, with blue representing closer regions and red representing farther regions.}
\label{distance map}
\end{figure}

\subsection{Target Prediction Module}
% 目标预测模块是 SONAR 范式中可学习的核心部分，使机器人能够从多通道语义地图中估计目标物体的空间坐标。该模块利用基于 U-Net 的架构，既能捕捉局部细粒度细节，又能获取全局上下文线索，以生成精确的目标位置。
The Target Prediction Module serves as the learnable core of the SONAR paradigm, enabling the robot to estimate the spatial coordinates of target objects from multi-channel semantic maps. Leveraging a U-Net based architecture, this module captures both local fine-grained details and global contextual cues to generate precise target positions.

% 该模块由一个多尺度编码器和一个坐标解码器组成。编码器采用一系列带有下采样操作的卷积块，从输入地图中提取分层语义特征。为了在保留关键语义信息的同时实现输入通道的降维，引入了一个 1×1卷积嵌入层：
The module consists of a multi-scale encoder and a coordinate decoder. The encoder employs a series of convolutional blocks with downsampling operations to extract hierarchical semantic features from the input maps. To achieve dimensionality reduction of the input channels while retaining essential semantic information, a $1\times1$ convolutional embedding layer is introduced:
% \begin{equation}
% \mathbf{F}_0 = \sigma \left(W_e * \mathbf{M}_{\text{in}} + b_e\right)
% \end{equation}
\begin{equation}
\mathbf{F}_0 = \sigma \left(W * M + b_e\right)
\end{equation}
where $M \in \mathbb{R}^{H \times W \times C}$ denotes the input semantic map, $W$ represents the embedding kernel, and $\sigma(\cdot)$ is a nonlinear activation. Subsequently, successive downsampling blocks are applied to progressively encode multi-scale spatial and semantic patterns, leading to a comprehensive feature representation $\mathbf{F}_L$.
\begin{equation}
\mathbf{F}_l = \phi_l\big(\mathrm{Down}(\mathbf{F}_{l-1})\big), \quad l = 1, \dots, L
\end{equation}
where $\phi_l(\cdot)$ denotes convolutional transformation at level $l$.

% 解码器从编码后的特征图中回归每个目标的二维坐标。它利用额外的卷积层来细化特征，随后通过池化层和全连接层来预测精确位置：
The decoder regresses the 2D coordinates of each target from the encoded feature maps. It utilizes additional convolutional layers to refine features, followed by pooling and fully connected layers to predict precise positions:
\begin{equation}
\hat{\mathbf{p}}_i = W_d \cdot \text{GAP}(\psi(\mathbf{F}L)) + b_d, \quad i=1,\dots,N_{\text{targets}}
\end{equation}
where $\psi(\cdot)$ denotes additional convolutional refinement, $\hat{\mathbf{p}}i \in \mathbb{R}^2$ is the predicted $(x,y)$ coordinate of the $i$-th target, and $\text{GAP}(\cdot)$ denotes global average pooling. The final output is structured as a tensor:
\begin{equation}
\hat{\mathbf{P}} = \left[\hat{\mathbf{p}}_1, \hat{\mathbf{p}}_2, \dots, \hat{\mathbf{p}}_{N_{\text{targets}}}\right] \in \mathbb{R}^{N_{\text{targets}} \times 2}
\end{equation}

The module is trained end-to-end by minimizing the mean squared error (MSE) between predicted and ground-truth coordinates:
\begin{equation}
\mathcal{L}{\text{target}} = \frac{1}{N_{\text{targets}}}\sum_{i=1}^{N_{\text{targets}}}\left|\hat{\mathbf{p}}_i - \mathbf{p}_i^{}\right|_2^2
\end{equation}
where $\mathbf{p}_i^{}$ is the ground truth coordinate of the $i$-th target.

% 通过持续整合实时语义观测数据，目标预测模块能够提供准确可靠的目标定位，这对于在复杂未知环境中的后续规划与导航至关重要。这种设计使得该模块能够在不同的目标类别和地图分辨率间实现通用化，支持灵活地整合到更广泛的声纳（SONAR）框架中，以实现自主且具备语义感知能力的导航。
By continuously integrating real-time semantic observations, the Target Prediction Module provides accurate and reliable target localization, which is critical for subsequent planning and navigation in complex, unknown environments. This design allows the module to generalize across different target categories and map resolutions, supporting flexible integration into the broader SONAR framework to enable autonomy with semantic awareness.

\begin{figure*}[t]
\centering
\includegraphics[width=6.5in]{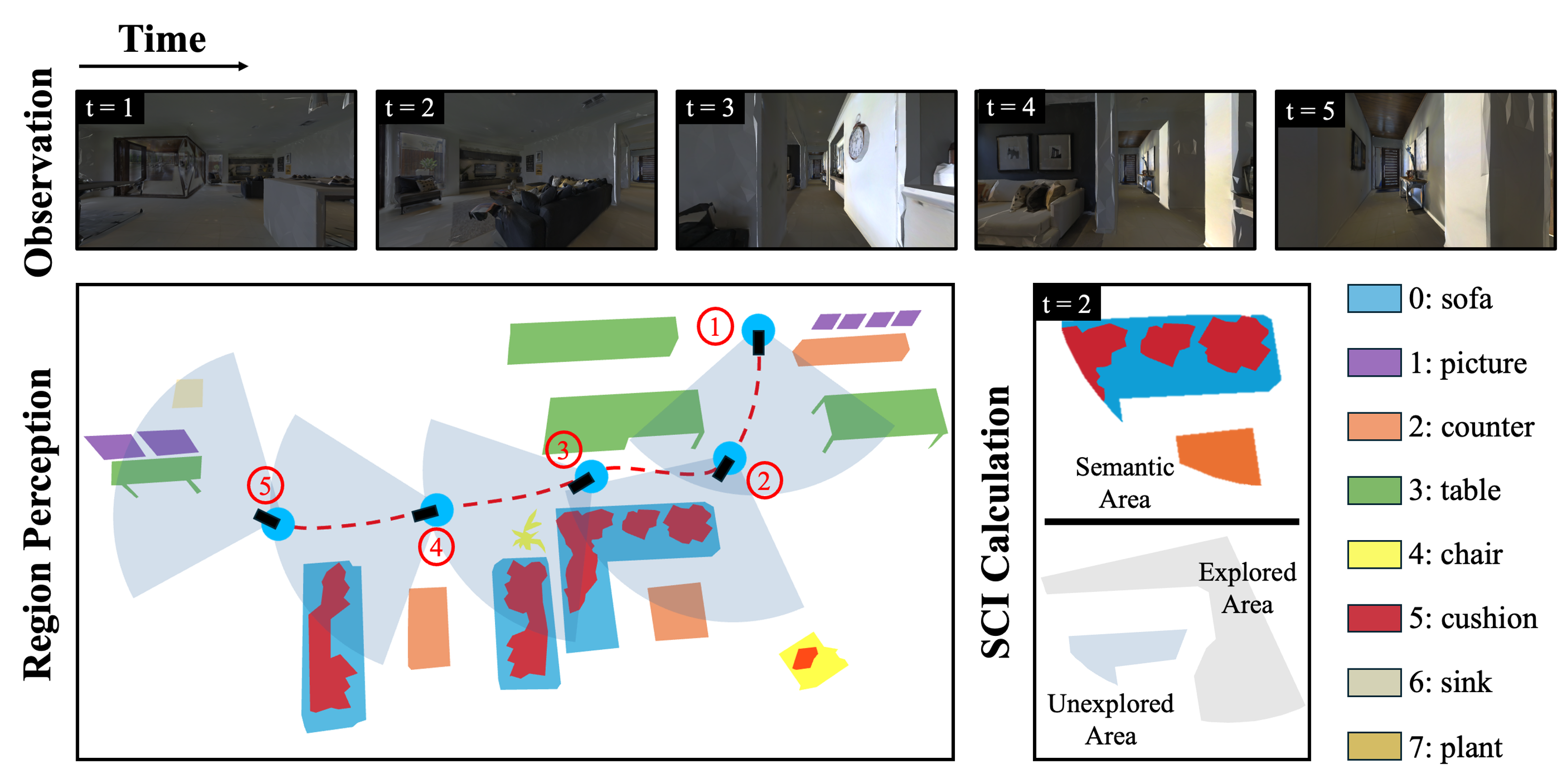}
\caption{An illustration of SCI calculation process. Taking the moment of t=2 as an example, the areas of the semantic area, explored area, and unexplored area within the current visual field observation of the agent are acquired. Subsequently, the value of SCI is determined based on the proportion of the semantic area.}
\label{sci}
\end{figure*}

\subsection{Fusion Exploration Strategy}
% 融合探索策略构成了 SONAR 实现动态自适应导航的核心逻辑。其主要目标是整合目标预测模块和基于视觉语言模型（VLM）的价值地图模块的输出，以便在具有不同语义线索密度的未知环境中持续选择最佳前沿点。通过利用多模态信息聚合与推理，该策略减轻了单模块方法的局限性，例如在语义稀疏区域鲁棒性降低，或在语义密集区域过度依赖局部信息等问题，从而提高探索效率和目标定位精度。
The Fusion Exploration Strategy constitutes the core logic of SONAR for achieving dynamically adaptive navigation. Its main objective is to integrate the outputs of the Target Prediction Module and the VLM based Value Map Module to consistently select optimal frontier points in unknown environments with varying semantic cue densities. By leveraging multimodal information aggregation and reasoning, this strategy mitigates the limitations of single-module approaches, such as reduced robustness in semantically sparse regions or over-reliance on local information in semantically dense areas, thereby improving both exploration efficiency and target localization accuracy. 

\textbf{1) Distance Map from Target Prediction:}
The predicted target coordinates $\hat{P} = [\hat{p}_1, \hat{p}_2, ..., \hat{p}_{N_{\text{targets}}}]$ are converted into a distance map $D$, where each cell represents the Euclidean distance to the nearest predicted target:
\begin{equation} 
D_i(x, y) = \sqrt{(x - x_i)^2 + (y - y_i)^2}, \quad i = 1, \dots, N
\end{equation}
\begin{equation}
D(x, y) = [D_1(x, y), D_2(x, y), \dots, D_{N}(x, y)]
\end{equation}

\textbf{2) Value Map Optimization from VLM:}
The Vision-Language Model module produces a semantic similarity score 
$S$ by matching features between real-time RGB images and target text prompts, and the value map $\mathbf{V}$ is updated using a field-of-view sector filling method. During the fusion exploration phase, spatial smoothing is applied to the value map by a $3\times3$ Gaussian filter, which removes fluctuations caused by local observation noise and ensures continuous and consistent values within the same semantically relevant region. The smoothed value map $\mathbf{V}_{\text{smoothed}}$ is normalized to the range $[0,1]$, with larger values indicating a higher probability that the region contains the target object.
\begin{align}
V'(x,y) &= \big( G_{3\times3} * V \big)(x,y) \\
V_{\text{smoothed}}(x,y) &= 
\frac{V'(x,y) - \min\limits_{(i,j) \in \Omega} V'(i,j)}
{\max\limits_{(i,j) \in \Omega} V'(i,j) - \min\limits_{(i,j) \in \Omega} V'(i,j)}
\end{align}
\noindent
where $V(x,y)$ is the original value map, $V'(x,y)$ is the smoothed map, $V_{\text{smoothed}}(x,y)$ is the normalized map, $G_{3\times3}$ is a $3\times3$ Gaussian filter for local smoothing, $*$ denotes convolution, $\Omega$ is the spatial domain of the map, and $(i,j)$ and $(x,y)$ indicate map cell coordinates.

\textbf{3) Semantic Cue Intensity (SCI) Calculation:}
The Semantic Cue Intensity (SCI) quantifies the semantic richness of the current environment, which is derived from the multi-target semantic map $SMap_{\text{multi}}$ and multi-target confidence map $CMap_{\text{multi}}$. Specifically, we iterate through all pixels within the robot current field of view, and calculate the proportion of pixels that satisfy both "having valid semantic labels (non-empty)" and "confidence $CMap_{\text{multi}}(x,y) > 0.6$" relative to the total number of pixels in the FOV. The mathematical expression of SCI is given by:
\begin{equation}
\text{SCI} = \frac{\sum_{(x,y) \in \text{FOV}} \mathbb{I}(S\neq\emptyset,\ C>0.6)}{N_{\text{FOV}}}
\end{equation}
where $\mathbb{I}(\cdot)$ denotes the indicator function (returning 1 if the condition inside is satisfied, and 0 otherwise). The value range of SCI is $[0, 1]$, and scenes are categorized based on SCI as follows:

- $\text{SCI} > 0.6$: Semantically dense scenes ;

- $0.3 \leq \text{SCI} \leq 0.6$: Moderately semantic ;

- $\text{SCI} < 0.3$: Semantically sparse scenes.

Let $w_{\text{pred}}$ represent the weight of the target prediction module, and $w_{\text{vlm}}$ represent the weight of the Vision-Language Model module. These two weights satisfy the constraint $w_{\text{pred}} + w_{\text{vlm}} = 1$. The rules for dynamic weight allocation are defined as:
\begin{equation}
\begin{cases} 
w_{\text{pred}} = 1 - \text{SCI}\\
w_{\text{vlm}} = \text{SCI} 
\end{cases}
\end{equation}
% where the first formula ensures a higher weight of the prediction module in semantically sparser scenes, and the second formula ensures a higher weight of the VLM module in semantically denser scenes.

\textbf{4) Dual-model Aggregation Reasoning (DAR):}
Based on the preprocessed distance map $D$, smoothed value map $V_{\text{smoothed}}$, and dynamic weights $w_{\text{pred}}$, $w_{\text{vlm}}$, we calculate the DAR score for each candidate exploration frontier point as a quantitative indicator of exploration priority.
For any candidate point $(x, y)$ in the frontier boundary map $F$, its DAR score is defined as:
\begin{equation}
DAR(x, y) = w_{\text{pred}} \cdot \frac{1}{D(x, y) + \epsilon} + w_{\text{vlm}} \cdot V(x, y)
\end{equation}
where $D(x, y)$ is the normalized distance map derived from predicted target coordinates, $V(x, y)$ is the smoothed value map obtained from the Vision-Language Model, $\epsilon$ is a small constant added to prevent division by zero, and $w_{\text{pred}}$ and $w_{\text{vlm}}$ are the corresponding fusion weights. Higher DAR values indicate regions that are considered more favorable for exploration.
Each candidate frontier point is evaluated using its DAR score, and the point with the highest score is chosen as the next target for exploration:
\begin{equation}
(x^*,y^*) = \arg\max_{(x,y)\in\mathbf{F}} DAR(x,y)
\end{equation}

\textbf{5) Real-time Multimodal Map Updates:}
Upon reaching $(x^*, y^*)$, new visual observations are acquired by the robot, which trigger several updates. Newly detected target and non-target objects are projected onto $SMap_{\text{target}}$ and $SMap_{\text{multi}}$, and the corresponding confidence maps $CMap_{\text{target}}$ and $CMap_{\text{multi}}$ are updated according to the specified equations. The SmolVLM module is invoked to compute the semantic score $s$ for the new observations, and the value map $\mathbf{V}$ is updated using the FOV fan-filling method. Finally, the updated semantic maps are input to the target prediction module, and the predicted target point set $\hat{\mathbf{P}}$ as well as the distance map $\mathbf{D}$ are regenerated.

\textbf{6) Target Locking Conditions and Termination Logic:}
The target object is considered reliably detected when cells with confidence values 
$CMap_{\text{target}}$ $>$ 0.7 appear in the Single-Target Semantic Map 
$SMap_{\text{target}}$. Candidate positions in the overlapping regions of the Single-Target Semantic Map and the Multi-Object Semantic Map are used to construct a fusion confidence map $CMap_{\text{fusion}}$, where the value of each cell is computed as the average of the corresponding confidences from the Single-Target Confidence Map and the Multi-Object Confidence Map. The cell with the maximum value in $CMap_{\text{fusion}}$ is selected as the preliminary target. A local neighborhood around this cell is used to compute a center position. The nearest navigable point within the explored area is chosen as the final navigation goal. Upon determination of this goal, the fusion exploration strategy terminates, and the robot follows the planned path to reach the target.

\subsection{Waypoint navigation}
Taking the target point generated in the Fusion Exploration Strategy as the global destination, this section employs the A* algorithm to complete global path planning. To achieve the synergy between global guidance and local control, the waypoint 1 meter away from the robot current position is first intercepted from the planned global path and used as the local navigation reference. Subsequently, the iPanner \cite{yang_iplanner_2023} local path planner is invoked, which takes this reference point as the local target to generate real-time motion control commands for the robot, ensuring the robot moves stably along the direction of the global path.

\begin{figure*}[t]
\centering
\includegraphics[width=6.5in]{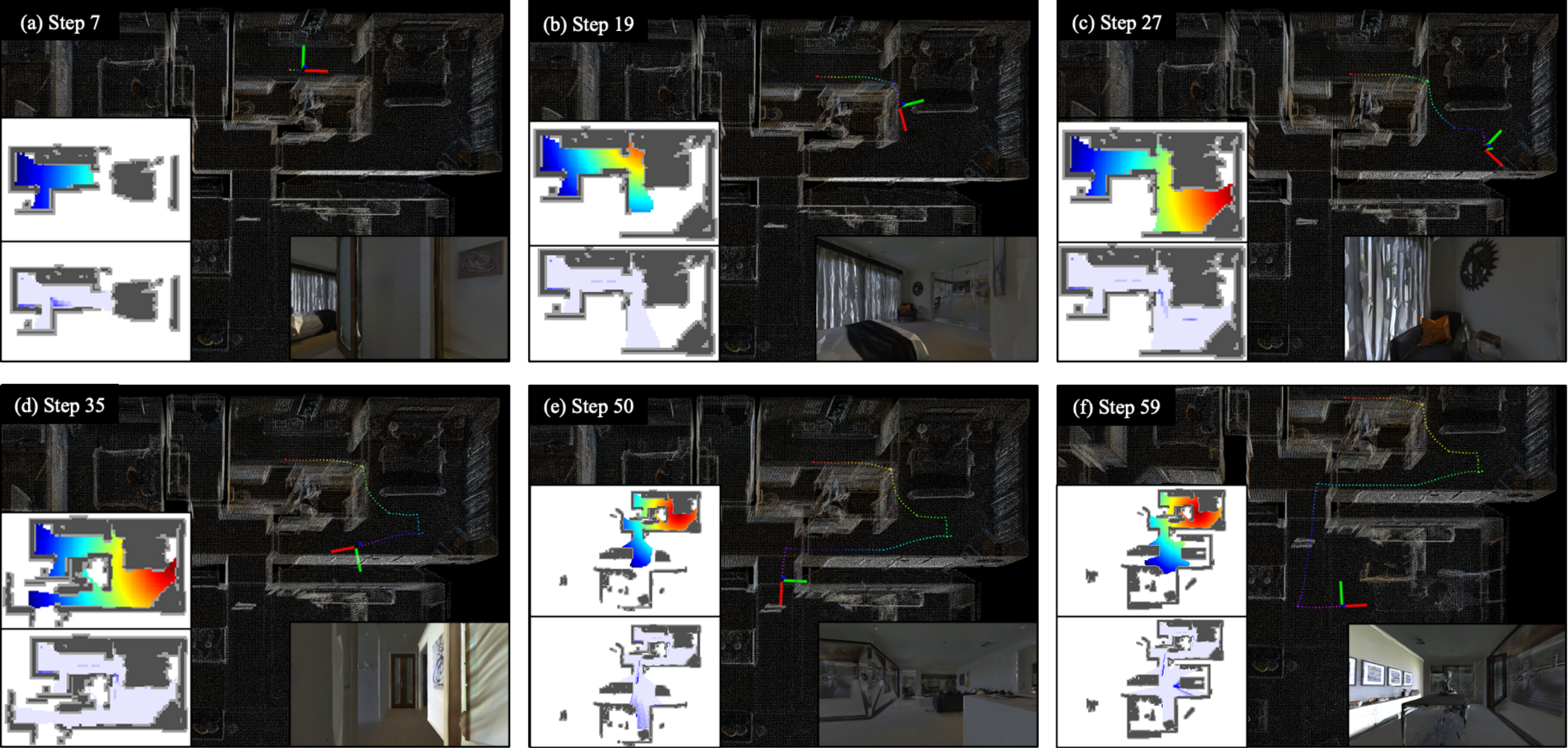}
\caption{Visualization of the navigation process of SONAR. We visualize parts of an ObjectNav episode on MP3D. For each step, we show the RGB image, the predicted target distance map, and the value map. At semantically dense moments (e.g., t = 50, 59), the agent switches its adaptive exploration strategy using the dynamic semantic cue intensity values.}
\label{shiyan}
\end{figure*}

\section{Experiments}
\label{sec:results}
\subsection{Experimental Setting}
\textbf{Dataset:} We evaluate the effectiveness and navigation efficiency of our proposed method on the widely used ObjectNav dataset MP3D in the Gazebo simulator. The MP3D validation dataset contains 11 indoor scenes, 21 object goal categories, and 2,195 object goal navigation episodes.

\textbf{Evaluation Metrics:} We use two metrics to evaluate algorithm performance: 1) Success Rate (SR): percentage of successful episodes; 2) Success weighted by Path Length (SPL): combines success rate and path efficiency. Higher values indicate better performance for all metrics.

\textbf{Implementation Details:} The agent operates within a simulation environment with a maximum of 500 steps per episode. It selects actions from a discrete set including MOVE FORWARD (0.25 m), TURN LEFT (30°), TURN RIGHT (30°) and STOP. The onboard camera is mounted at a height of 0.88 m with a 79° horizontal field of view, processing 640×480 RGB-D images. For perception and reasoning tasks, the system employs a combination of advanced models: GroundingDINO for object detection, with Mobile-SAM utilized for semantic segmentation to perform additional semantic processing. The value map is generated by SmolVLM-500M.

\textbf{Baselines:}
% 现有方法根据其对监督的依赖程度和零样本泛化能力分为三类。对于非零样本方法，有监督方法（如 SemExp [4] 和 PONI [17]）和无监督方法（如 ZSON [13]）都被考虑在内。然而，这些方法对未见类别的泛化能力有限。在零样本类别中，大多数方法是无监督的，并结合语义推理或语言模型来指导探索。例如，ESC [30] 和 L3MVN [26] 使用语义地图与大语言模型相结合来选择前沿点，而 VLFM [25] 则通过视觉语言模型生成价值地图，引导智能体朝向目标。
Existing methods are categorized into three groups according to their dependence on supervision and zero-shot generalization ability. For non-zero-shot methods, both supervised approaches, such as SemExp \cite{chaplot2020object} and PONI \cite{ramakrishnan2022poni}, and unsupervised approaches, such as ZSON \cite{majumdar2022zson}, are considered. These methods, however, exhibit limited generalization to unseen categories. In the zero-shot group, most methods are unsupervised and incorporate semantic reasoning or language models to guide exploration. For instance, ESC \cite{zhou2023esc} and L3MVN \cite{yu2023l3mvn} use semantic maps combined with large language models to select frontier points, while VLFM \cite{yokoyama2024vlfm} generates value maps via visual-language model to direct the agent toward the target.

\begin{table}[t]
\centering
\caption{Comparison with previous work on MP3D.}
\begin{tabular}{lcccc}
\toprule
Method & Unsupervised & Zero-shot & SR $\uparrow$ & SPL $\uparrow$ \\
\midrule
SemEXP \cite{chaplot2020object} & No & No & 36.0 & 14.4 \\
PONI \cite{ramakrishnan2022poni} & No & No & 31.8 & 12.1 \\
\midrule
ZSON \cite{majumdar2022zson} & Yes & No & 15.3 & 4.8 \\
\midrule
CoW \cite{gadre2023cows} & Yes & Yes & 7.4 & 3.7 \\
ESC \cite{zhou2023esc} & Yes & Yes & 28.7 & 14.2 \\
L3MVN \cite{yu2023l3mvn} & Yes & Yes & 34.9 & 14.5 \\
VLFM \cite{yokoyama2024vlfm} & Yes & Yes & 36.4 & 17.5 \\
\midrule
SONAR (Ours) & No & Yes & 38.4 & 17.7 \\
SONAR w/o VM & No & No & 36.2 & 16.4 \\
SONAR w/o TPM & Yes & Yes & 35.6 & 15.3 \\
\bottomrule
\end{tabular}
\label{table:comparison}
\end{table}

\subsection{Experimental Results}
% 在 MP3D 数据集上的实验结果如表1所示。与现有方法相比，SONAR 取得了最佳表现，其 SR 为 38.4%，SPL 为 17.7%。与VLFM相比成功率sr提高了5.5%，spl提高了1.2%，与次优的SemExp相比成功率sr提高了6.7%，spl提高了22.9%。与监督方法相比，SONAR 在没有VLM Module的情况下相比于 SemExp 成功率提高了0.6%，spl提高了6.3% 。与次优的poni相比成功率提高了13.8%，spl提高了37.2%。与零样本方法相比，SONAR 在没有预测地图的情况下与次优的L3MVN相比成功率提高了2.0%，spl提高了5.5%。
The experimental results on the MP3D dataset are presented in Table~\ref{table:comparison}. Compared with existing methods, SONAR achieves the best performance with an SR of 38.4\% and an SPL of 17.7\%. Specifically, SONAR improves SR by 5.5\% and SPL by 1.2\% over VLFM, and surpasses the second-best method SemEXP by 6.7\% in SR and 22.9\% in SPL, demonstrating its superiority in both success rate and navigation efficiency.
Compared with supervised methods, SONAR without the VLM module achieves a 0.6\% higher success rate and a 6.3\% higher SPL than SemEXP, and improves the success rate by 13.8\% and SPL by 37.2\% over PONI, demonstrating its efficient exploration capability.
Compared with Zero-shot methods, SONAR without the prediction map also achieves a 2.0\% higher success rate and a 5.5\% higher SPL than the suboptimal method L3MVN.

\subsection{Ablation Study}
% 目标的定位：1.object map 2.semantic map 3.object & semantic map
% 模块：1.prediction 2.VLM
% 如表 \ref {table:Ablation} 所示，我们对四个组件进行了系统的消融实验，包括单目标定位（STL）、多目标定位（MOL）、目标预测图（TPM）和价值图（VM），以严格评估我们关键模块的有效性。
% STL 专注于对目标物体进行映射和定位，为单目标导航提供必要的空间信息。相比之下，MOL 为所有检测到的物体构建地图和定位，捕捉环境内的全面语义分布。TPM 利用实时语义地图预测潜在的目标位置，为探索提供前瞻性指导。VM 从当前观测中生成局部价值图，提供动态线索以支持明智的导航决策。
As presented in Table \ref{table:Ablation}, systematic ablation experiments were conducted on the four components, including Single-Target Localization (STL), Multi-Object Localization (MOL), Target Predicted Map (TPM) and Value Map (VM), to rigorously evaluate the effectiveness of our key modules. 
STL focuses on mapping and localizing the target object, providing essential spatial information for single-target navigation. In contrast, MOL constructs maps and localizations for all detected objects, capturing the comprehensive semantic distribution within the environment. TPM leverages the real-time semantic map to predict potential target locations, enabling anticipatory guidance for exploration. VM generates a local value map from current observations, providing dynamic cues to support informed navigation decisions.

% 移除了MOL模块的sr为35.5，spl为16.5.
% 移除了STL模块的sr为36.6，spl为16.8.
% 移除了VM模块的sr为36.2，spl为16.4.
% 移除了TPM模块的sr为35.6，spl为15.3.

% 为了进一步研究每个组件的作用，我们考虑了各组件间的组合。完整配置下的成功率（SR）为 38.4，路径长度加权成功率（SPL）为 17.7，作为参考基准。
% 移除多目标定位（MOL）模块：成功率降至 35.5，路径长度加权成功率降至 16.5。这种性能下降表明，全局语义上下文在增强导航鲁棒性方面起着至关重要的作用。没有多目标定位模块，系统仅依赖单目标线索，导致对环境的感知不够全面。
% 移除单目标定位（STL）模块：系统的成功率为 36.6，路径长度加权成功率为 16.8，与完整模型相比下降幅度最小。这表明多目标定位模块可以通过提供多个物体的语义地图部分补偿单目标定位模块的缺失。然而，与完整模型的差距证实了精确的单目标定位仍具有独特价值。
% 移除价值地图（VM）模块：性能下降至成功率为 36.2，路径长度加权成功率为 16.4。这一下降突出了基于动态价值的引导对于优化局部导航决策的重要性，特别是在模糊或杂乱区域。
% 移除目标预测模型（TPM）模块：成功率和路径长度加权成功率分别降至 35.6 和 15.3，这是所有设置中下降幅度最大的。这一发现强调了预测目标建模的重要性，因为目标预测模型提供的预测线索能显著提高探索效率。
% 总体而言，结果表明所有组件都有积极贡献，但它们的相对重要性有所不同。目标预测模型的影响最大，证实了其作为关键预测机制的作用。多目标定位模块和价值地图模块也通过丰富语义上下文和动态决策做出了重大贡献，而单目标定位模块仍然是提供准确位置线索的基础。这些模块之间的协同作用解释了为什么完整系统能实现最佳的整体性能。
To systematically evaluate the importance of each component, we conducted an ablation study based on different component combinations. In the full configuration, the model achieves an SR of 38.4 and an SPL of 17.7, which serves as the reference for subsequent analysis.

Removing MOL: SR drops to 35.5 and SPL to 16.5, indicating that global semantic context is crucial for robust navigation. Without MOL, the system relies solely on single-target cues. This results in less comprehensive environmental awareness.

Removing STL: SR drops to 36.6 and SPL to 16.8, representing the smallest decline compared to the full model. This suggests that MOL can partially compensate for STL by providing semantic maps of multiple objects. However, the gap to the full model confirms that precise single-target localization still offers unique benefits.

Removing VM: SR drops to 36.2 and SPL drops to 16.4. This highlights the importance of dynamic, value-based guidance for refining local navigation decisions, especially in ambiguous or cluttered regions.

Removing TPM: SR drops to 35.6 and SPL drops to 15.3, representing the most significant decrease among all components. This underscores the critical role of predictive target modeling. TPM provides anticipatory cues that substantially enhance exploration efficiency.

Overall, the results show that all components contribute positively, but their relative importance differs. TPM has the largest impact, confirming its role as a critical predictive mechanism. MOL and VM also make substantial contributions by enriching semantic context and dynamic decision-making, while STL remains fundamental for providing accurate localization cues. The synergy among these modules explains why the complete system achieves the best overall performance.

\renewcommand{\arraystretch}{1.4}
\begin{table}[t]
\centering
\caption{Ablation study of different components of SONAR.}
\begin{tabular}{cccccc}
\hline
\multicolumn{2}{c}{Localization Method} & \multicolumn{2}{c}{Mapping Method} & \multirow{2}*{SR$\uparrow$} & \multirow{2}*{SPL$\uparrow$}   \\
\cmidrule(lr){1-2} \cmidrule(lr){3-4}
STL & MOL  & TPM & VM    \\

\hline
\checkmark  &            & \checkmark &            & 33.1 & 14.8 \\
\checkmark  &            &            & \checkmark & 32.1 & 13.4 \\
\checkmark  &            & \checkmark & \checkmark & 35.5 & 16.5 \\

            & \checkmark & \checkmark &            & 34.7 & 15.3 \\
            & \checkmark &            & \checkmark & 33.5 & 14.2 \\
            & \checkmark & \checkmark & \checkmark & 36.6 & 16.8 \\

\checkmark  & \checkmark & \checkmark &            & 36.2 & 16.4 \\
\checkmark  & \checkmark &            & \checkmark & 35.6 & 15.3 \\
\checkmark  & \checkmark & \checkmark & \checkmark & 38.4 & 17.7 \\

\hline
\end{tabular}
\label{table:Ablation}
\end{table}

\section{Conclusions}
\label{sec:conclusion}
% 在本文中，我们介绍了 SONAR，这是一种在跨模态推理范式下进行聚合推理的语义目标导航框架。SONAR 融入了一种动态自适应探索机制，该机制将基于语义地图的目标预测模块与由视觉 - 语言模型驱动的价值地图模块相结合。这种设计使智能体能够根据环境中的语义线索强度自适应地平衡这两个模块的贡献。具体而言，在语义稀疏的场景中，SONAR 依靠全局语义推理来定位目标，而在语义密集的环境中，它利用视觉 - 语言对齐来提高探索精度。此外，我们提出了一种新颖的融合策略，将多尺度语义地图与置信度地图相结合。通过利用互补的多源信息，这种方法能够准确推断目标位置，减轻因局部观测限制而产生的误差，并显著减少导航过程中的冲动决策错误。
In this paper, we introduce SONAR, a semantic object navigation framework with aggregated reasoning under a cross-modal inference paradigm. SONAR incorporates a dynamically adaptive exploration mechanism that integrates a semantic map-based target prediction module with a Vision-Language Model driven value map module. This design enables the agent to adaptively balance the contributions of the two modules according to the semantic cue intensity in the environment. Specifically, in semantically sparse scenarios, SONAR relies on global semantic reasoning to locate the target, whereas in semantically dense environments, it exploits Vision-Language alignment to enhance exploration accuracy. Furthermore, we propose a novel fusion strategy that combines multi-scale semantic maps with confidence maps. By leveraging complementary multi-source information, this approach allows for accurate inference of the target position, mitigates errors arising from the limitations of local observations, and significantly reduces impulsive decision-making errors during navigation.\\

\noindent\textbf{Acknowledgements}
This work is supported by National Natural Science Foundation of China under Grant 62473191, Shenzhen Key Laboratory of Robotics Perception and Intelligence (ZDSYS20200810171800001), Shenzhen Science and Technology Program under Grant 20231115141459001, RCBS20221008093305007, Guangdong Basic and Applied Basic Research Foundation under Grant 2025A1515012998, Young Elite Scientists Sponsorship Program by CAST under Grant 2023QNRC001, and the High level of special funds (G03034K003) from Southern University of Science and Technology, Shenzhen, China.

% \section*{Declarations}

% \noindent\textbf{Data Availability} 
% The datasets generated and analysed during this study are available from the corresponding author on reasonable request.

    \bibliographystyle{IEEEtran}	
    \bibliography{SONAR}

\end{document}